\documentclass[letterpaper, 10 pt, journal, twoside]{IEEEtran}
\IEEEoverridecommandlockouts

\usepackage[pdftex]{graphicx}
\graphicspath{{../pdf/}{../jpeg/}}
\DeclareGraphicsExtensions{.pdf,.jpeg,.png,.eps}
\usepackage{amsmath}
\usepackage{amssymb}
\usepackage{algorithmic}
\usepackage{array}
\usepackage{stfloats}
\usepackage{url}
\usepackage{booktabs}
\usepackage[nolist,nohyperlinks]{acronym}
\usepackage{multicol}
\usepackage[bookmarks=true]{hyperref}
\usepackage{tabularx}
\usepackage{bm}

\usepackage{xfrac}
\usepackage{cleveref}

\newcommand{\threeD}{$3D$}
\newcommand{\twoD}{$2D$}
\newcommand{\Rthree}{$\mathbb{R}^{3}$}
\newcommand{\manifold}{\mathcal{M}}
\newcommand{\manifoldtwo}{\mathfrak{m}}
\newcommand{\meshmanifold}{\mathcal{M}_{\triangle}}

\newcommand{\meshmanifoldtwo}{\mathfrak{m}_{\triangle}}

\newcommand{\mapping}{\mathcal{H}}
\newcommand{\jacobian}{J_{\meshmanifoldtwo \leftarrow \meshmanifold}}
\newcommand{\polperp}{\mathcal{P}^{\perp}_{\meshmanifoldtwo}}
\newcommand{\polsurf}{\mathcal{P}^{\rightarrow}_{\meshmanifoldtwo}}
\newcommand{\surf}{\rightarrow}
\newcommand{\matrixthree}{\in \mathbb{R}^{3 \times 3}}

\acrodef{RMP}{\emph{riemannian motion policy}}
\acrodefplural{RMP}{\emph{riemannian motion policies}}
\acrodef{NDT}{\emph{non-destructive testing}}
\acrodef{UAV}{\emph{unmanned aerial vehicle}}
\acrodef{DOF}{\emph{degrees of freedom}}
\acrodef{MAV}{\emph{micro aerial vehicle}}
\acrodef{OMAV}{\emph{omnidirectional micro aerial vehicle}}
\acrodef{COM}{\emph{center of mass}}
\acrodef{TOF}{\emph{time of flight}}
\acrodef{ASIC}{\emph{axis-selective impedance control}}
\acrodef{VIO}{\emph{visual inertial odometry}}
\acrodef{IMU}{\emph{inertial measurement unit}}
\acrodef{PD}{\emph{proportional-derivative}}
\acrodef{PI}{\emph{proportional-integral}}
\acrodef{NURB}{\emph{Non-uniform rational B-Spline}}
\acrodef{ROS}{\emph{Robot Operating System}}
\acrodef{CGAL}{\emph{The Computational Geometry Algorithms Library}}
\acrodef{TSDF}{\emph{Truncated Signed Distance Field}}

\begin{document}

\title{Mesh Manifold based Riemannian Motion Planning for Omnidirectional Micro Aerial Vehicles}
\author{Michael Pantic$^{1}$, Lionel Ott$^{1}$, Cesar Cadena$^{1}$, Roland Siegwart$^{1}$, and Juan Nieto$^{1,2}$
\thanks{Manuscript received: October, 15, 2021; Revised January, 11, 2021; Accepted February, 6, 2021.}%
\thanks{This paper was recommended for publication by Editor Nancy Amato upon evaluation of the Associate Editor and Reviewers’ comments.}%
\thanks{This work was supported by funding from the National Center of Competence in Research (NCCR) on Digital Fabrication, and NCCR Robotics through the Swiss National Science Foundation.}
\thanks{$^{1}$ The authors are with Autonomous Systems Lab, ETH Z\"urich, 8092 Z\"urich, Switzerland.
{(e-mail: mpantic@ethz.ch; lioott@ethz.ch; cesarcadena.lerma@gmail.com;
rsiegwart@ethz.ch;  juannieto@microsoft.com)}}%
\thanks{$^{2}$ Author is with Mixed Reality \& AI Lab, Microsoft, 8001 Z\"urich, Switzerland, work conducted when the author was with ETH Z\"urich. }
\thanks{Digital Object Identifier (DOI): see top of this page.}}%
\markboth{IEEE Robotics and Automation Letters. Preprint Version. Accepted February, 2021}
{Pantic \MakeLowercase{\textit{et al.}}: Mesh Manifold based Riemannian Motion Planning for Omnidirectional Micro Aerial Vehicles}
\maketitle%
%
%
%
\begin{abstract}
This paper presents a novel on-line path planning method that enables aerial robots to interact with surfaces.
We present a solution to the problem of finding trajectories that drive a robot towards a surface and move along it.
Triangular meshes are used as a surface map representation that is free of fixed discretization and allows for very large workspaces.
We propose to leverage planar parametrization methods to obtain a lower-dimensional topologically equivalent representation of the original surface.
Furthermore, we interpret the original surface and its lower-dimensional representation as manifold approximations that allow the use of Riemannian Motion Policies (RMPs), resulting in an efficient, versatile, and elegant motion generation framework. We compare against several Rapidly-exploring Random Tree (RRT) planners, a customized CHOMP variant, and the discrete geodesic algorithm. Using extensive simulations on real-world data we show that the proposed planner can reliably plan high-quality near-optimal trajectories at minimal computational cost. The accompanying multimedia attachment demonstrates feasibility on a real OMAV. The obtained paths show less than $10 \%$ deviation from the theoretical optimum while facilitating reactive re-planning at kHz refresh rates, enabling flying robots to perform motion planning for interaction with complex surfaces.
\end{abstract}
\IEEEpeerreviewmaketitle
\section{Introduction}
\IEEEPARstart{R}{ecent} developments in \acp{MAV} that can interact with and exert forces on the environment make a variety of new use cases such as contact inspection, spraying, and painting, to name a few, possible. The advent of \acp{OMAV} allows these interactions to take place in arbitrary orientations \cite{bodie2020active}, which was previously impossible with traditional \acp{MAV} due to their underactuated nature.
Current planning methods \cite{oleynikovaOpenSourceSystemVisionBased2018} often use a discretized map such as an octree or a voxel grid to store occupancy information and plan collision-free trajectories using sampling-based or optimization-based algorithms.
There is a large range of literature about planning through unknown spaces in continuously updating discretized maps while avoiding obstacles. In these use cases, the exact spatial location of the trajectory is of less importance as long as it remains collision-free.
\begin{figure}[]
    \centering
    \includegraphics[width=\columnwidth]{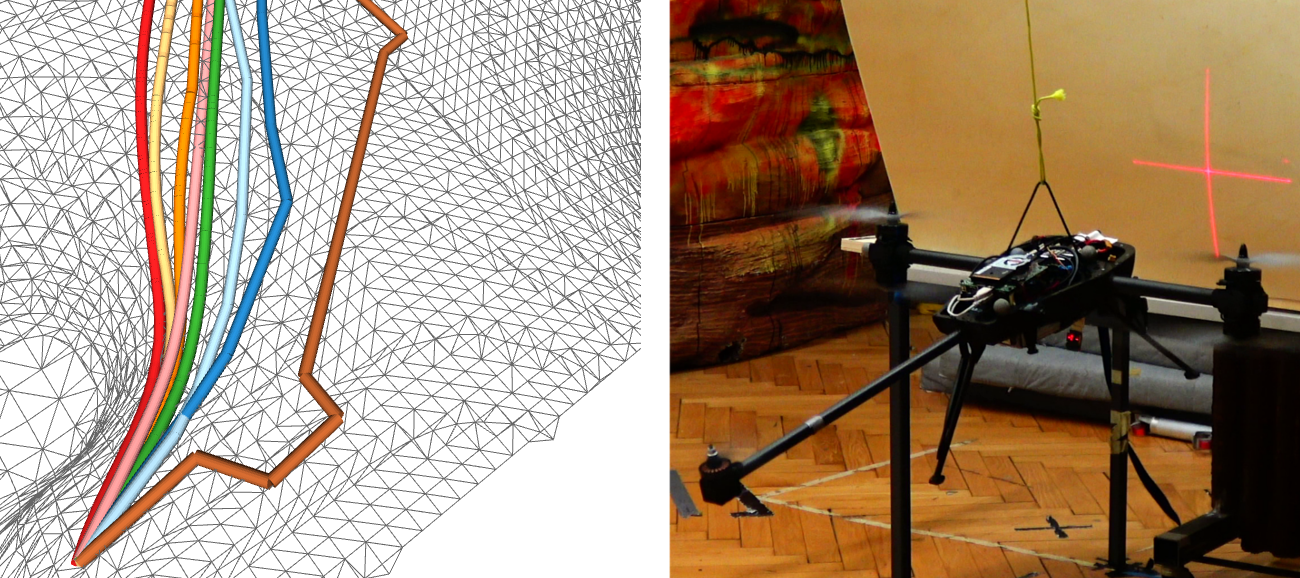}
    \caption{Left: Illustration of a typical planning run with all planners on the \texttt{hilo} scenario. Red is the proposed, green the DGEO, and brown the RRT*-Con planner. Blue shades correspond to the RRT*-Sam, yellow shades to the RRT*-Pro, and pink to CHOMP. Right: Example of an OMAV inspecting a surface using the proposed planner.}
    \label{fig:planner_ilustration}
    \vspace{-4mm}
\end{figure}
Using these existing methods for aerial interaction of surfaces with an \ac{OMAV} can be cumbersome and ill-posed as the requirements are completely different. One problem is the scalability of map representations currently used for \ac{MAV} motion planning. Three dimensional fixed-discretization map representations do not scale well in terms of resolution or map size.
\ac{OMAV}s however are able to perform centimeter-level manipulation tasks on very large a priori known workspaces that can be hundreds of meters in size, giving rise to the need for a very efficient high-resolution surface representation and an associated planning method. In most industrial, inspection, and outdoor use-cases, maps can be obtained up-front, which makes on-line mapping less important.
Another important challenge is the trajectory generation itself, as aerial interaction takes place in relation to a surface rather than free space. 
Planning a high-precision path along a surface should exploit connectedness of the surface, which implicit representations such as \acp{TSDF} are unable to do.
Furthermore, the determination and planning of the relative orientation of an \ac{OMAV} with respect to a surface is challenging in the presence of gimbal lock and singularities. These problems can be mitigated by using a surface map that contains a notion of orientation within the surface that is always valid.
\subsection{Related Work}
Many planning approaches designed specifically for \acp{MAV} use a smooth spline representation of the trajectory and exploit their differential flatness. Typical examples are spline motion primitives \cite{muellerComputationallyEfficientMotion2015} and optimization-based spline planning \cite{richter2016polynomial}. 
These methods generally assume that the \ac{MAV} moves in free-space. When obstacles are to be avoided, optimization-based techniques that use collision gradients such as CHOMP \cite{ratliffCHOMPGradientOptimization2009} are popular. Similarly, collision gradients are used in \cite{oleynikovaOpenSourceSystemVisionBased2018} to optimize \ac{MAV} trajectory splines in order to obtain safe free-space trajectories.
Another option is sampling-based planner, such as rapidly-exploring random trees (RRT) \cite{lavalle2001randomized} and its variants (e.g. RRT* \cite{karaman2011sampling}, RRT-Connect \cite{844730}), to find obstacle-free paths. However, trajectory planning for aerial inspection and manipulation should not just avoid obstacles but actively follow surfaces and objects. Sampling-based planners can be used for this by constraining the sampling space. \cite{tognon2018control} uses an RRT planner that samples in a task space constrained by all admissible end-effector states. Another option is to sample unconstrained but project onto the closest admissible space \cite{kingston2018sampling}.
While obstacle avoidance can allow inaccuracies as long as they are safe, surface following or interaction quality is more dependent on the surface representation quality and resolution. Implicit surface representations store occupancy information that allows the recovery of the surface. Common examples are octrees \cite{hornungOctoMapEfficientProbabilistic2013} and hash-based voxel grids \cite{oleynikovaVoxbloxIncremental3D2017}.
While fixed-discretized representations are very efficient for retrieval and lookup of data, their practical scalability in terms of resolution and workspace is constrained by memory usage.
Explicit surface representations directly store the boundary between occupied and free space. For 3D applications, common techniques are manifold splines \cite{guManifoldSplines2006}, \acp{NURB} \cite{pieglNURBSBook1997} and triangular meshes. A major advantage of these representations is their geometric nature which does not rely on a fixed discretization or a fixed resolution and allows efficient exploitation of geometric neighborhood connectedness.
Surfels \cite{schopsSurfelMeshingOnlineSurfelBased2018} and point clouds are often used as a surface representation that is easy to obtain from sensor data. However, they do not encode surface connectivity, which could be exploited by planning algorithms.
 \cite{bircherStructuralInspectionPath2015} uses triangular meshes for inspection planning but does not exploit connectedness, and instead samples states based on triangle normals and centers. Similarly, \cite{ruetzOVPCMesh3D2019} uses meshes to represent free space without relying on fixed discretization and to extract traversability information, without explicitly using the connected nature of the mesh. The interpretation of a surface mesh as an approximation of a manifold yields a natural mathematical parametrization of the connectedness. A typical example is the calculation of discrete geodesics \cite{xinImprovingChenHan2009}, which are defined as the shortest path between two points on a mesh manifold. The work in \cite{doi:10.1177/0278364919891775} introduces trajectory optimization on Riemannian Manifolds for obstacle-avoidance and field-of-view aware planning for \acp{MAV} but uses analytically represented manifolds.
Another class of planners that exploit the manifold structure of task and configuration space are \acfp{RMP} \cite{ratliffRiemannianMotionPolicies2018}. By optimally combining multiple motion policies across manifolds, \acp{RMP} enable the formulation of planning and control problems in arbitrary Riemannian manifolds as long as a Jacobian that relates them locally is obtainable.

\subsection{Contributions}
In this paper, we propose a solution to the problem of efficiently planning trajectories that follow and/or approach a-priori known surfaces.
We use triangular meshes as a surface representation that is not limited by discretization or resolution. To make use of this representation for interaction planning, we propose a motion generation algorithm that is based on differential geometry principles and Riemannian motion policies \cite{ratliffRiemannianMotionPolicies2018}.
Our contributions are the following:
\begin{itemize}
    \item The formulation of an efficient $2D-3D$ mesh manifold parametrization for planning.
    \item A highly efficient path planning framework based on mesh manifolds and Riemannian motion policies.
    \item Extensive experimental verification in simulation and comparison to a variety of sampling-based and optimization-based planning algorithms for surface following.
\end{itemize}
While in this paper we concentrate on the surface following use-case for \acp{OMAV}, the proposed approach of using meshes as an approximation to a mathematical manifold for RMP-based planning can be generalized to a variety of other applications, such as explicitly encoding nullspace manifolds.
\subsection{High-level system overview}
The concept of the proposed planner is easiest understood by a well-known analogy from cartography. In our daily lives we generally do not care about the spherical shape of our world. Instead, we use flattened, cartesian representations to plan a trajectory to a desired goal. 
Analogous, we flatten the \threeD{} mesh surface map to a lower-dimensional representation using a suitable parametrization function. We then exploit the topological equivalence (\textit{homeomorphism}) between the original surface and the flattened representation to induce acceleration fields as Riemannian motion policies that generate the desired trajectories along and towards the surface in \threeD{}. Thereby we use the explicit surface encoding of a triangular mesh efficiently.
In the remainder of this paper we detail the manifold approximation using meshes (\Cref{sec:meshmanifold}) and present an \ac{RMP} planner using this approximation (\Cref{sec:rmpplanner}). We show extensive evaluations (\Cref{sec:experiments}) and discuss the results (\Cref{sec:conclusion}).
\section{Mesh Manifolds}
\label{sec:meshmanifold}
The defining property of our framework is the use of conventional triangular meshes as a computationally tractable approximation to a smooth Riemannian manifold that represents a surface in \Rthree. By using the mesh representation which is inherently geometric and has no fixed discretization, our system is not constrained to a specific resolution or extent of the map.
In this section, we lay out the theoretical background of making the surface connectedness of a mesh easily accessible to the planner. We do so by interpreting the mesh as an approximation to a manifold embedded in a higher-dimensional space and obtaining a flattened, axis-aligned homeomorphic representation where the two axes of \textit{``going along the surface''} are perpendicular to the remaining axis of moving \textit{``towards or away''} of the surface. Such a mapping also provides explicit orientation information on the surface, as the angle with respect to the axes along the surface is always defined.
To ensure the correctness of the proposed planner, we show that the approximations made are valid and additionally give strategies for computationally efficient implementation.
\subsection{Prerequisities and definitions}
In the following, $\manifold$ denotes a surface embedded in $\mathbb{R}^{d}$. Here, we refer to the mathematical definition of a surface - a $(d-1)$-dimensional manifold.
We assume $\manifold$ to be smooth, free of holes, have a defined boundary, and be homeomorphic to a disc (no self-intersections). We require $\manifold$ to be Riemannian, i.e. to have a smoothly varying positive-definite inner product on the tangent space at every point. While the smoothness criterion is needed for a sound theoretical base, the proposed algorithm works relatively well on not perfectly smooth geometries as demonstrated in the results section.
Without loss of generality, we assume $\manifold$ to be embedded in $\mathbb{R}^{3}$ in the following. 
The homeomorphic \twoD\ representation of $\manifold$ is subsequently denoted by $\manifoldtwo$. 
Intuitively, we use the flattened representation $\manifoldtwo$ of the \threeD\ surface $\manifold$, together with a one-to-one mapping between them, as a \twoD\ coordinate representation that is embedded in the surface. We use triangular meshes as a surface representation that approximates the properties of an ideal manifold. To distinguish the ideal, mathematical manifold from its approximate triangular mesh implementation, we subsequently refer to the mesh as $\meshmanifold$ respectively $\meshmanifoldtwo$. 
\subsection{Notation}
We denote a point $i$ that is part of the \threeD\ mesh $\meshmanifold$ as $P_{i} = (x_{i}, y_{i}, z_{i}) \in \mathbb{R}^{3}$. Similarly, a point $j$ on the \twoD\ mesh $\meshmanifoldtwo$ is denoted by $p_{j} = (u_{j}, v_{j}) \in  \mathbb{R}^{2}$. It is important to note that $P_{i}$, respectively $p_{j}$, refer to an arbitrary point that is part of the mesh surface. Vertices are a subset of these points and in the following referred to as $\overline{P}_{i}$, respectively $\overline{p}_{j}$.
The mesh structure is formalized as the set of all vertices $V(\meshmanifold)=\{\overline{P}_{0}, ..., \overline{P}_{N}\}$, respectively $V(\meshmanifoldtwo) = \{\overline{p}_{0},...,\overline{p}_{n}\}$.
A triangle formed by vertices and edges is formalized as a ordered triplet of vertices $T_{ijk} = \{\overline{P}_{i}, \overline{P}_{j}, \overline{P}_{k}\}$ on the \threeD\ mesh, respectively  $t_{ijk} = \{\overline{p}_{i}, \overline{p}_{j}, \overline{p}_{k}\}$ on the \twoD\ mesh. Any arbitrary point on the mesh is part of at least one triangle.

\subsection{Coordinate mapping}
\begin{figure}
    \centering
    \includegraphics[width=0.4\textwidth]{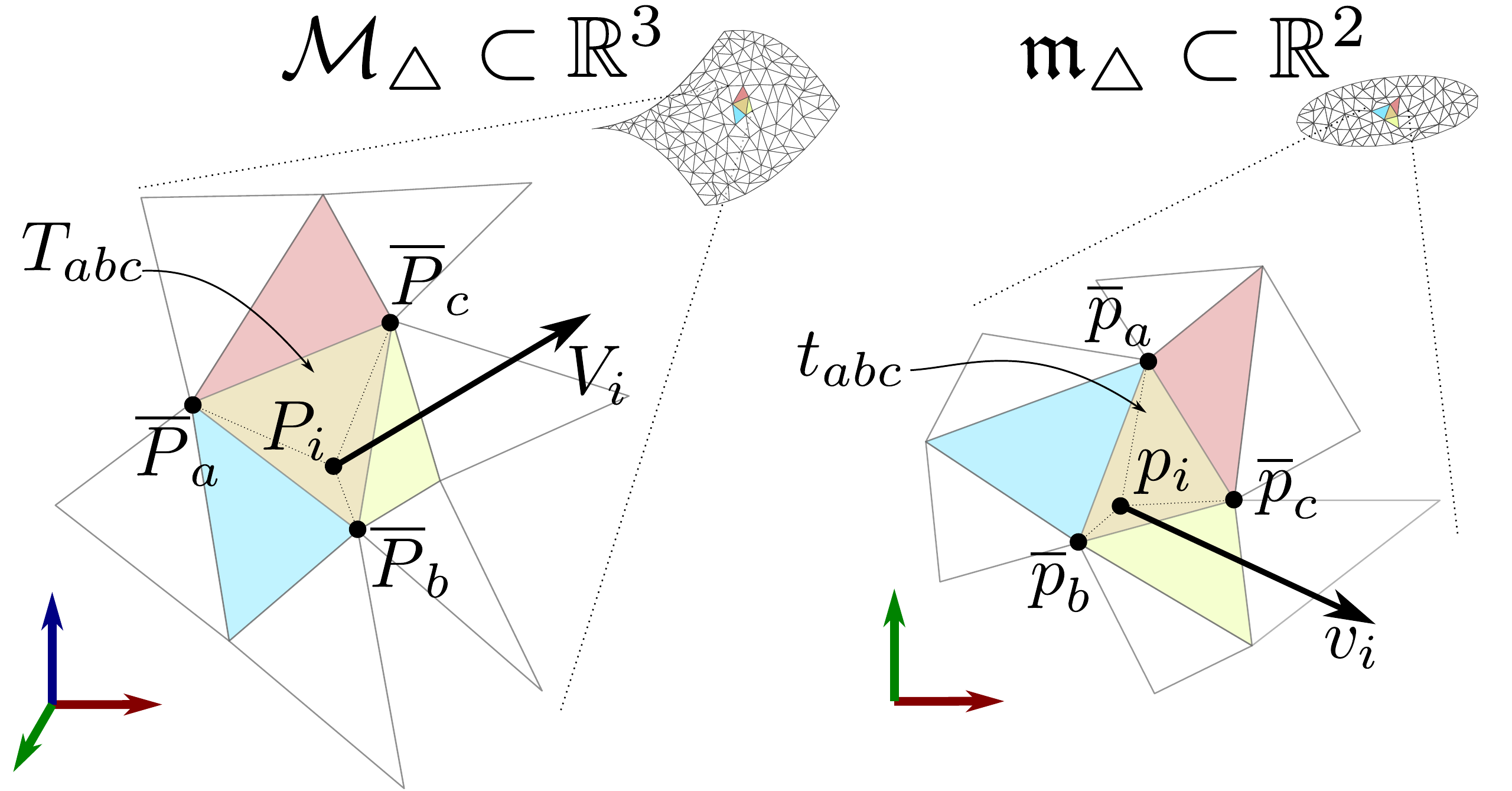}
    \caption{Closeup of a pair of triangles on both mesh manifolds that shows the used notation. The left triangles are three dimensional while the right triangles are two dimensional. The two points $P_{i}$ and $p_{i}$ have the same barycentric position w.r.t. the orange triangle. The vectors $V_{i}$ and $v_{i}$ are related to each other by the $\jacobian$.}
    \label{fig:triangle}
    \vspace{-4mm}
\end{figure}

We obtain the \twoD{} representation $\meshmanifoldtwo$ of $\meshmanifold$ by applying a parametrization $\mapping$ that maps each \threeD\ vertex to a \twoD\ vertex while retaining topology:
\begin{align}
    \overline{p}_{k} &= \mapping(\overline{P}_{k})\\
    \overline{p}_{k} & \in V(\meshmanifoldtwo) \subset \mathbb{R}^{2},\; \overline{P}_{k} \in V(\meshmanifold) \subset \mathbb{R}^{3}, \nonumber
\end{align}
We assume $\mapping$ to be a one-to-one mapping and to generate a valid triangulation $\meshmanifoldtwo$ that is topologically equivalent to $\meshmanifold$. To obtain a mapping of an arbitrary point $p \in \meshmanifoldtwo$ to their corresponding point $P \in \meshmanifold$ and vice versa, we use standard barycentric coordinates \cite{mobius1827barycentrische}.
Assuming $p \in \meshmanifoldtwo$ lies in triangle $t_{ijk}$ we obtain the corresponding point $P \in \meshmanifold$ by first calculating the barycentric coordinates $\mathcal{B}(p,t_{ijk}) =\{\beta_{1}, \beta_{2}, \beta_{3}\},\; \beta_{1..3} \in \mathbb{R}$
of $p$ relative to \twoD{} triangle $t_{ijk}$ and then applying these barycentric coordinates to the corresponding \threeD{} triangle $T_{ijk}$:
\begin{equation}
	P \in \meshmanifold = \Gamma(p,t_{ijk},T_{ijk}) = \mathcal{B}^{-1}(\mathcal{B}(p,t_{ijk}),T_{ijk})
    \label{eq:relation_triangle}
\end{equation}
Note: $\Gamma$ is a short-hand notation for the full mapping defined by $\mathcal{B}$ and $\mathcal{B}^{-1}$.
Figure \ref{fig:triangle} visualizes the mapping process.
Additionally to the \twoD-\threeD\ coordinate mapping we augment the \twoD\ coordinates of $\meshmanifoldtwo$ with a third dimension $h$ that is normal to the surface. For any triangle on $\meshmanifold$, this dimension coincides with the normal of that triangle.

\subsection{Selection of mapping function}
One important property needed for this planner is a surface flattening method that provides a bijective one-to-one mapping $\mapping$. Algorithms such as \textit{Tutte Barycentric Mapping} \cite{tutte1963draw}, \textit{Discrete Authalic Parametrization} \cite{desbrunParametrization} and \textit{Floater Mean Value Coordinates} \cite{floater2003mean} provide such a mapping. All three parametrization can generate \twoD\ homeomorphic discs with either square or circular boundaries.
Furthermore, as in regular navigation a conformal (angle-preserving) mapping is strongly preferred as otherwise mappings of velocities and directions between \twoD\ and \threeD\ can be inaccurate. Only the \textit{Floater Mean Value Coordinates} explicitly optimize for low angular distortion and are therefore chosen as our default implementation of $\mapping$. We use circular border parametrizations, as planners operating on square borders showed divergent behavior for points on the boundaries. A more detailed overview of the different methods can be found in \cite{cgal:parametrization}.

\subsection{Implementation}
To facilitate efficient and fast translation of arbitrary points between $\meshmanifold$ and $\meshmanifoldtwo$, the result of $\mapping$ is computed for every vertex $\overline{P}$ on startup and cached in forward and reverse hashmaps. This allows translation of vertices with a computational complexity of $\mathcal{O}(1)$. 
In order to obtain the closest triangle to an arbitrary point, we use accelerated queries in axis-aligned bounding-box (AABB) \cite{cgal:aabb} trees for both $\meshmanifold$ and $\meshmanifoldtwo$.
\section{Planning Framework}
\label{sec:rmpplanner}
In this section, we explain how the previously described mesh manifolds are used to plan paths relative to a surface. Our framework is based on Riemannian Motion Policies (\acp{RMP}) \cite{ratliffRiemannianMotionPolicies2018}. The proposed planner is efficient and well defined due to two properties that follow from the mesh manifold approximations. Firstly, by exploiting the \twoD\ nature of a surface embedded in \threeD\ space we effectively reduce the dimensionality of the problem and thus mitigate scaling effects in terms of computational complexity and memory usage. 
Secondly, having a gravity-independent and singularity-free orientation on the surface allows the construction of a valid orientation of the OMAV at any point in the surface. 

\subsection{Riemannian Motion Policies}
Riemannian Motion Policies (\acp{RMP}) \cite{ratliffRiemannianMotionPolicies2018} provide a framework to formulate and combine multiple motion policies on different manifolds. In the following, a brief summary is given. Without loss of generality, we assume to have only two different manifolds, configuration space $\mathcal{Q}$ and task space $\mathcal{X}$. We denote positions and their derivatives in configuration space as $q, \dot{q}, \Ddot{q}$ and similarly for the task space $x, \dot{x}, \Ddot{x}$.
We convert between the two spaces using a task-map $\phi(q) = x$.
Additionally, a position-dependent analytic Jacobian 
$J_{\mathcal{X} \leftarrow \mathcal{Q}} = \frac{\partial \phi}{\partial q}$
that maps velocities and accelerations \textit{locally} between the two spaces is needed. 
A Riemannian Motion Policy $\mathcal{P}_{\mathcal{X}}$ in the task space is defined as the tuple $(f, A)_{\mathcal{X}}$, where $f$ is an acceleration function $f(x,\dot{x})$ and $A(x,\dot{x})$ the smoothly varying, positive semidefinite Riemannian metric associated with the policy. 
As shown in \cite{ratliffRiemannianMotionPolicies2018}, two policies can be summed as a metric-weighted average 
\begin{equation}
    \mathcal{P}^{a}_{\mathcal{X}} + \mathcal{P}^{b}_{\mathcal{X}}  = ((A^{a} + A^{b})^{+}(A^{a}f^{a} + A^{b}f^{b}), A^{a}+A^{b})_{\mathcal{X}}
\end{equation} to provide an optimal solution for the combined system. This property is very powerful, as it allows the decomposition of complex problems into many simple policies that can be combined optimally.

Furthermore, to transform a policy from one space to another, the pullback operator is used:
\begin{equation}
    \textit{pull}_{\mathcal{Q}}((f,A)_{\mathcal{X}}) = ((J^{T}AJ)^{+}J^{T}Af, J^{T}AJ)_{\mathcal{Q}}.
\end{equation}
This effectively gives us an acceleration in $\mathcal{Q}$ that executes a policy (or combination thereof) defined in taskspace $\mathcal{X}$.
\subsection{Mesh Manifold as Task Space}
We use $\meshmanifold$ in $\mathbb{R}^3$ as the configuration space $\mathcal{Q}$ and $\meshmanifoldtwo$ as the task space  $\mathcal{X}$. 
While in many applications, such as dexterous manipulation, the configuration space has a higher dimensionality than the task space, here we exploit the space transformation capabilities of \acp{RMP} to simplify the \threeD{} planning problem and exploit surface connectedness. Due to the induced coordinate mapping on the surface, it is straightforward to generate spatiotemporal trajectories that follow the induced acceleration field on the surface and map them into $\mathbb{R}^3$.
We use the coordinate mapping defined in the previous section as task map and obtain the needed Jacobian $\jacobian \in \mathbb{R}^{3 \times 3}$ analytically by exploiting the bijective triangle mapping between $\meshmanifold$ and $\meshmanifoldtwo$ and the mapping of barycentric coordinates for arbitrary points on the two meshes.
By taking the partial derivatives of eq. \ref{eq:relation_triangle} for all dimensions, we obtain the first 2 columns of $\jacobian$. The last column follows from the definition of the $h$ axis and corresponds to the normalized normal of the 3D triangle.
For a point $P \in \meshmanifold$ that is an element of triangle $T$ with the corresponding triangle $t \in \meshmanifoldtwo$, we obtain the Jacobian as follows:
\begin{equation}
J_{\meshmanifoldtwo \leftarrow \meshmanifold}(P) = 
\begin{bmatrix}
\frac{\partial}{\partial x}\Gamma(P, t, T)\ \frac{\partial}{\partial y}\Gamma(P, t, T)\ \mathcal{N}(T) \\
\end{bmatrix}
\end{equation}
where $\mathcal{N}(\cdot)$ returns the normalized normal vector for a triangle $T$.
The analytical derivation of $\jacobian$ is constant for all points $P$ on a specific triangle $T$.
This allows us to formulate policies that follow and approach $\meshmanifoldtwo$, but execute them in \threeD{} on an \ac{OMAV} that operates on $\meshmanifold$ embedded in $\mathbb{R}^{3}$.
\subsection{Surface Attractor Policy}
We decouple the surface following problem into two independent policies formulated on $\meshmanifoldtwo$. The first policy, subsequently called $\polperp$, drives the trajectory onto the surface of $\meshmanifoldtwo$. This policy corresponds to the generic attractor policy in \cite{ratliffRiemannianMotionPolicies2018} and is defined as
\begin{align}
	\polperp &= (f^{\perp},A^{\perp}) \\
	\label{eq:polperp}
	f^{\perp} &= \alpha^{\perp} \cdot \mathbb{S}(\textbf{0} - p_{0}) - \beta^{\perp} \dot{p}_{0} = \Ddot{p_{0}} \\
	A^{\perp} &= \text{diag}(0,0,1) \matrixthree,
\end{align}
\noindent where $\alpha^{\perp}$ and $\beta^{\perp}$ are tuning parameters, and $\mathbb{S}$ is the soft-normalization function
\begin{equation}
    \mathbb{S}(z) =\frac{z}{|z| + \gamma\ log(1+exp(\gamma |z|))}
\end{equation} with tuning parameter $\gamma$ defined in \cite{ratliffRiemannianMotionPolicies2018}, $p_{0} \in \meshmanifoldtwo$ is the current position w.r.t. $\meshmanifoldtwo$ and $\dot{p}_{0}$ the current velocity. $\polperp$ induces an acceleration field in $\meshmanifoldtwo$ that points towards the surface. 
By pulling the policy from task space to configuration space using metric $A^{\perp}$, we constrain the policy to be only acting on the dimension perpendicular to the surface. The resulting acceleration field in $\meshmanifold$ respectively \Rthree\ smoothly drives the trajectory to a desired surface distance (usually $0$).
\subsection{Surface following policy}
The second policy, $\polsurf$ drives the trajetory to a desired position $p_{des} = [u_{des}, v_{des}, h_{des}]$ on the surface. Due to the metric only $u,v$ affect the execution of the policy.
\begin{align}
	\polsurf &= (f^{\surf},A^{\surf}) \\
	\label{eq:polsurf}
	f^{\surf} &= \alpha^{\surf} \cdot \mathbb{S}(p_{des} - p_{0}) - \beta^{\surf} \dot{p}_{0} = \Ddot{p_{0}} \\
	A^{\surf} &= \text{diag}(1,1,0) \matrixthree,
\end{align}
The policy $\polsurf$, when pulled to the configuration space $\meshmanifold$ with metric $A^{\surf}$, induces an acceleration field that follows the surface and drives the trajectory to the specified goal position.
\subsection{Parameter Tuning}
For each policy a parameter set $\mathcal{T} = \{\alpha, \beta, \gamma\}$ is needed.
By varying $\alpha^{\perp}, \beta^{\perp}$ against $\alpha^{\surf}$ and $\beta^{\surf}$, the relative strength and aggressiveness of the two policies can be chosen. This has an especially large influence on the trajectory planning from free-space towards a goal on the surface. Depending on the relative strength, the contact with the surface is sought as soon as possible, gradually or towards the end of the trajectory. For all experiments we used the following values: $\mathcal{T}^{\surf}=\{0.7, 13.6, 0.4\}, \mathcal{T}^{\perp}=\{20.0,30.0, 0.01\}$. The tuning process is relatively intuitive, stable, and fast, as the results of a re-tuning can be visualized in real-time.
\subsection{Orientation and Offset}
At any location on $\meshmanifoldtwo$ we can trivially determine the $u$ and $v$ axis direction, as in any regular Cartesian coordinate system. This is a side benefit of using a planar parametrization of an explicit surface representation. By mapping these directions onto  $\meshmanifold$ we obtain an on-surface orientation that is always defined, regardless of the actual \threeD{} surface orientation.
By calculating the normalized inverse of the Jacobian $\jacobian$ at a specific point $P$, we obtain the on-surface orientation as column vectors. 
As an example, we derive the orientation $R$ for aligning an $\ac{OMAV}$ body $x$-axis with direction tangent to the surface along $u$, and the body $z$-axis with the direction perpendicular to the surface.
The resulting rotation matrix $R \in \mathbb{R}^{3}$ is therefore constructed as
\begin{equation}
    R = \begin{bmatrix}
    -\jacobian^{-1}[:,1]^{T} \\  
    \jacobian^{-1}[:,1] \times \jacobian^{-1}[:,3]^{T} \\ \jacobian^{-1}[:,3]^{T}
    \end{bmatrix}^{T},
    \label{eq:orientation}
\end{equation}
where $[:,i]$ selects the $i$-th 1-based column.
Due to the induced distortion by mapping $\mapping$, the \threeD{} representation of the $u$ and $v$ axis are not necessarily orthogonal, therefore we only use the $u$ axis and the normal and obtain the third direction by cross-product (second row in \cref{eq:orientation}).
The mesh-based planning method applies identically for meshes constructed based on the original meshes. One use-case would be the inflation of the mesh with methods such as the one presented in \cite{5570949} to guarantee a certain distance from the surface without self-intersections.
\section{Experiments}
\label{sec:experiments}
\begin{figure}[ht]
    \centering
    \includegraphics[width=\columnwidth]{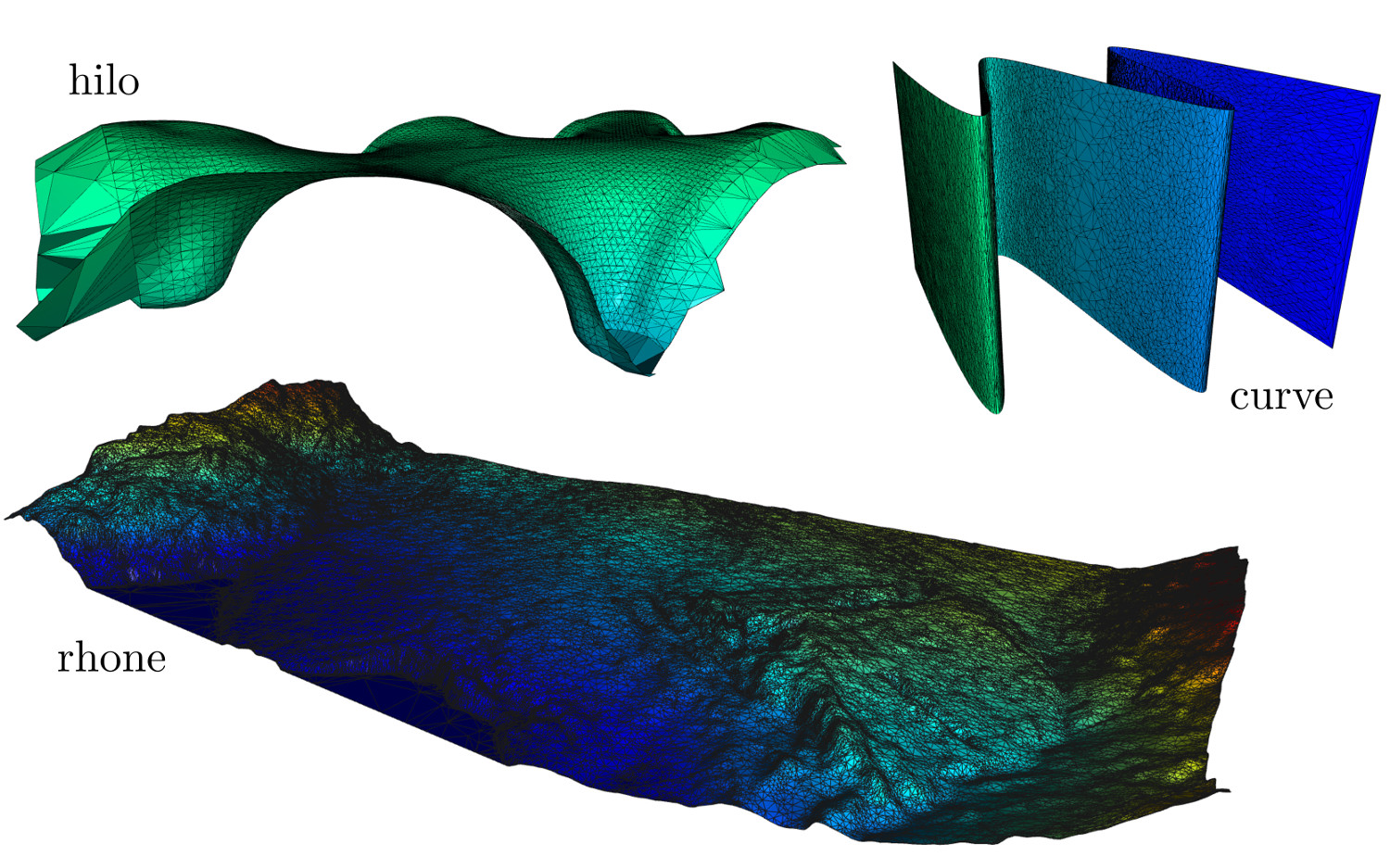}
    \caption{Rendering of all three scenarios, not to scale. \texttt{hilo} originates from a laser scan of a thin-shelled concrete roof and \texttt{rhone} is generated from high-res digital elevation maps of a glacier. \texttt{curve} is artificially generated and serves to demonstrate sampling bottlenecks and planning difficulties in highly variable geometry under ideal conditions.}
    \label{fig:meshes}
\end{figure}

We evaluate the proposed planning method in three different scenarios, shown in \cref{fig:meshes} and \cref{tab:scenario_details}.
For each scenario $100$ planning tasks are generated by randomly sampling a start and end location on the surface and solved by each planner variant. The planner performances are evaluated in simulation, and feasibility demonstrated on a real OMAV\footnote{In the written paper we concentrate on the simulation experiments as the bare execution of the planned trajectory mostly measures controller performance and not planning quality. See accompanying multimedia attachment for videos of the OMAV experiments.}.
\subsection{Comparison Planners}
We compare against a broad range of geometric, sampling-based, and optimization-based planners. All compared methods solve the same problem: to find the shortest path between two points while being constrained to the surface.
As a geometric planner, we use the widely used theoretically-optimal (shortest path) discrete geodesic algorithm (subsequently abbreviated as DGEO) proposed in \cite{xinImprovingChenHan2009} and implemented in CGAL \cite{cgal:shortest_path}. Three variants of RRT* implemented using OMPL \cite{sucan2012the-open-motion-planning-library} represent sampling-based planners, while CHOMP \cite{ratliffCHOMPGradientOptimization2009} is used as an optimization-based planner\footnote{To facilitate fair comparison, we use state-of-the-art C++ implementations with as little changes as possible, usually limited to adjustments in the cost function implementation to enable all planners to use the same mesh mapping back-end e.g. for distance lookups.}.
\begin{table}
\centering
\begin{tabular}{lrcrr}
\hline
\quad & Faces  & Extent $[m]$ & Surface $[m^{2}]$ & Setup $[s]$ \\
 \hline
 rhone & $126.4$k & $625.9 \times 463.1 \times 88.6$ & $160802.1$ & 2.31        \\
 curve & $20.0$k  & $37.5 \times 35.2 \times 25.0$   & $4234.1$ & 0.18  \\
 hilo  & $16.1$k  & $21.2 \times 9.4 \times 5.2$     & $209.6$  & 0.13      \\
\hline
\end{tabular}
\caption{Metric sizes and meshing details of each scenario.}
\label{tab:scenario_details}
\vspace{-8mm}
\end{table}
The variants of the RRT* algorithm differ in how they randomly generate new planning states (here, positions in \threeD\ space).
The \textit{sampling} variant (called RRT*-Sam), leverages the explicit surface representation of the mesh to uniformly sample positions directly on the mesh surface. Transitions between states are considered valid if paths between positions are within $1$ cm of the surface, verified at 5\% steps along the path. This is a rather loose verification but needed to clear sampling bottlenecks (discussed in the next section). RRT(*) is often used with task-space constrained samplers \cite{tognon2018control}. Here we use it as an example to show the behavior of constrained sampling in difficult geometries, as any extension needs to clear the same sampling bottlenecks.
The \textit{projecting} variant \cite{kingston2018sampling}, (called RRT*-Pro) mimics planning using an implicit surface representation. Therefore, any direct addressing of the surface is impossible and sampling cannot be performed on the surface. Instead, states are randomly sampled in the encompassing \threeD\ volume and are then projected onto the surface using a Jacobian. For RRT*-Pro all paths between states are considered valid, even if they leave the surface. However, the resulting path is smoothed and again projected back onto the surface using the Jacobian. Note that RRT*-Pro is evaluated on analytic geometries in \cite{kingston2018sampling}, which makes geometric lookups and Jacobian calculation more efficient, whereas here it suffers from more expensive operations on a mesh map.
Both, RRT*-Sam and RRT*-Pro use a fixed time budget and return the shortest path found within the allocated time or a failure state. For both variants we evaluate a time budget of $1$ s and $0.25$ s, which are indicated by ``$1$'' respectively ``$\sfrac{1}{4}$''.
In contrast, the \textit{connecting} RRT (called RRT*-Con) starts sampling from start and goal independently and terminates as soon as a valid connection is found or the allocated time (1 s) has passed.
The CHOMP planner uses an adjusted potential function, as its published version is used to avoid obstacles. We observed almost impossible convergence by using a potential function without slack around the ideal state (= path on the surface). Therefore, we define zero potential if within $0.1$ m of the surface, quadratic within $1.1$ m and linear otherwise. 
The parameters and configuration for all used planners are constant across all scenarios.
Our evaluation additionally serves to provide insight into the underlying nature of the surface following problem and to show that the ability to exploit an explicit surface representation \textit{and} its connectedness is highly beneficial.
\Cref{fig:planner_ilustration} shows an example trajectory for each planner.
\subsection{Success Rate}
\begin{figure}[h!]
    \centering
    \includegraphics[width=\columnwidth]{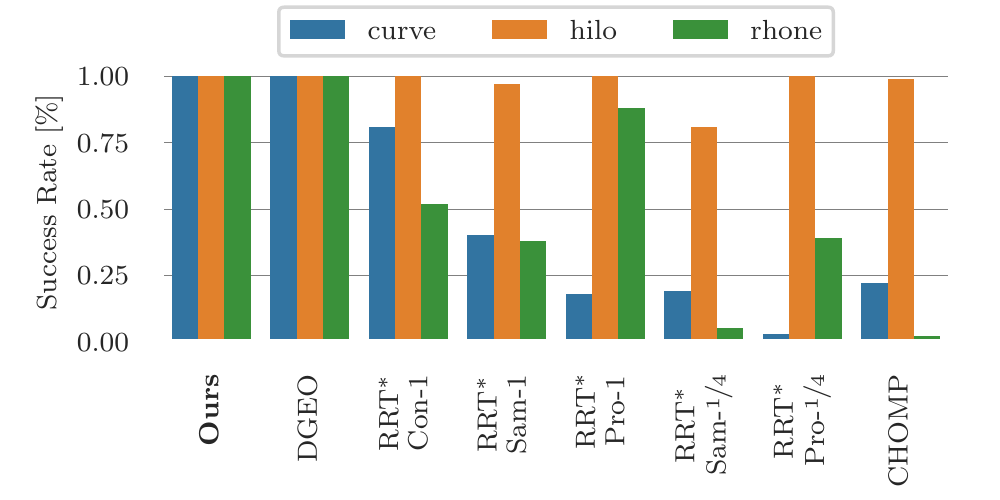}
    \vspace*{-8mm}
    \caption{Success rates as percentages for each planner. The color indicating the scenario is the same for all subsequent plots.}
    \label{fig:success_rate}
    \vspace{-2mm}
\end{figure}

The success rates for all 8 tested planners are plotted in \cref{fig:success_rate}. For the RMP planner, the success criterion is fulfilled if the trajectory is within $0.5$ cm of the desired target and at rest. The discrete geodesic planner is guaranteed to converge exactly and the sampling-based planners are considered successful if an exact connected solution is returned.
The proposed planner successfully converged for all trajectories. The sampling-based planners were relatively successful on the moderate geometry of scenario \texttt{hilo}. The variants that are able to exploit the explicit surface connectedness (\textit{Con/Pro}) consistently outperformed the projecting planner on the difficult geometry of scenario \texttt{curve}. This can be attributed to the non-uniform sampling w.r.t. the surface in the \textit{RRT*-Sam} variants. CHOMP worked well on the mid-sized, easy scenario \texttt{hilo} but got stuck in local minima often (\texttt{curve}) or would need to be re-tuned to account for the largest map (\texttt{rhone}).
On all subsequent plots, only successful trajectories are shown. Note that this can introduce biases in the data, as e.g. all successfully planned paths for some of the RRT variants on scenario \texttt{curve} are strictly within the planar parts of the surface and do not go across one of the bends.
\subsection{Planning duration}
\begin{figure}[!h]
    \centering
    \includegraphics[width=\columnwidth]{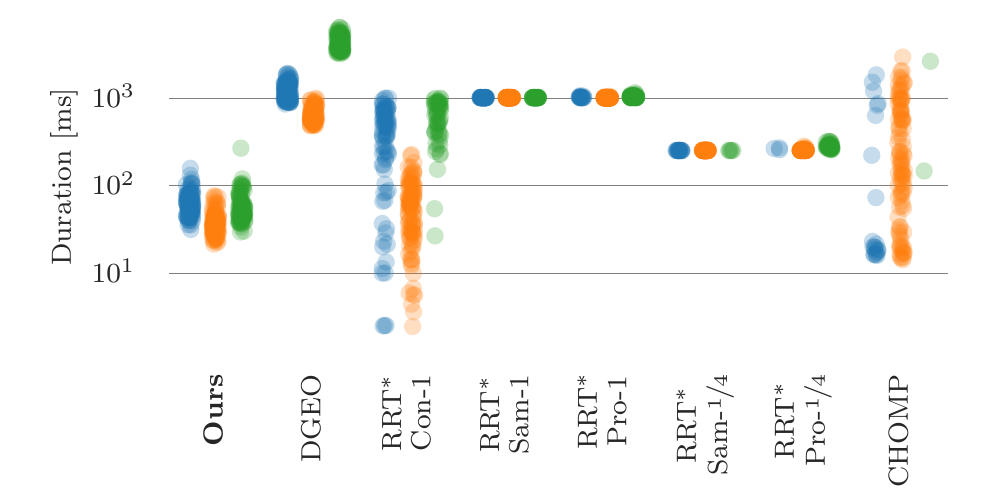}
    \vspace*{-8mm}
    \caption{Average duration for planning of a full trajectory from start to goal. Note the logarithmic scale.}
    \label{fig:timings}
    \vspace{-2mm}
\end{figure}

Another important metric is the duration to reach a successful planning state. 
For the fixed-time variants of RRT* the duration is constant within timer resolution. For all RRT-based planners as well as the discrete geodesic algorithm, only the actual solving time (including path interpolation) is counted and setup times are excluded. The execution time of CHOMP highly depends on the scenario and task to solve. The results are displayed in \cref{fig:timings}.
The RMP planner by itself only generates acceleration fields. In order to facilitate a fair comparison, we measure the time it takes to evaluate this acceleration field at a temporal resolution of $100$ Hz and integrate the acceleration and velocity using a trapezoidal integrator to obtain the full trajectory. 
For execution on a real robot, this would not be necessary as we can simply evaluate the policies at the current position and velocity to follow the trajectory at any time. Additionally, the time to obtain the mapping $\mapping$ once on startup is indicated in \cref{tab:scenario_details} for each scenario. Over all results presented here, the RMP planner needed on the order of just $\bm{10\ \mu s}$ per such iteration, which enables seamless re-planning at rates of $>10$ kHz. This enables smooth replanning for semi-manual steering on the surface, reactive planning with onboard sensors, or reacting to controller deviations.

\subsection{Smoothness}
\begin{figure}[h!]
    \centering
    \includegraphics[width=\columnwidth]{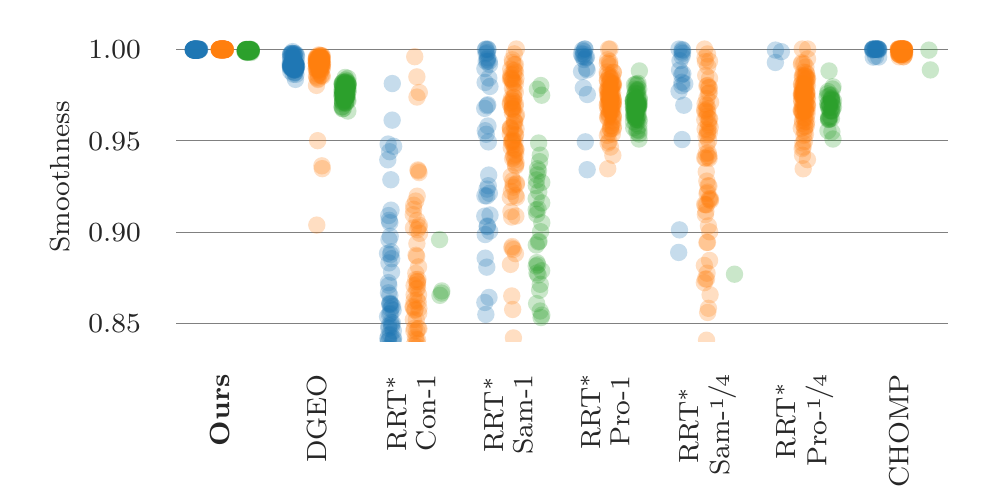}
    \vspace*{-8mm}
    \caption{Trajectory smoothness for all planner-scenario combinations. The plot is truncated at a smoothness of $0.85$ - there are values below this. The differential geodesic algorithm provides a lower bound of smoothness of the mesh, as it follows the mesh exactly at all times.}
    \label{fig:smoothness}
    \vspace{-4mm}
\end{figure}

For execution on the real robot, path smoothness is a desired property as changes in direction should be followed gradually and not in a jerky motion.
Here we evaluate the three dimensional angular similarity between subsequent segments of a trajectory. Formally, the angular similarity $\rho$ between two vectors $A,B$ is defined as
\begin{equation}
	\rho = 1 - \frac{1}{\pi}\ cos^{-1}\left(\frac{A \cdot B}{|A||B|}\right),
\end{equation}
and the smoothness as the average $\rho$ over a full trajectory.
Values very close to $1.0$ indicate very little angular changes and short average segments, whereas values below $0.95$ become visibly jagged. $0.5$ corresponds to an average change of angle of $90$ degrees.
\Cref{fig:smoothness} visualizes the trajectory smoothness over all evaluated planners. The RRT-based planners produce on average a lot less smooth trajectories than the proposed planner. Especially RRT-Connect results in quite jagged trajectories, as it terminates on the first found connecting path. The very smooth trajectories produced by the proposed planners can largely be attributed to their physical nature and high sampling rate. As the planner produces an acceleration field that is integrated, the trajectory must always change gradually (at least on a very local scale). Paths obtained by RRT variants could be smoothed and post-processed further e.g. by fitting splines. However, this is not part of the planner per se and could add arbitrary increases of runtime and impairment of accuracy. CHOMP is forced to generate smooth paths due to the smoothness cost used in the optimization.
\subsection{Surface Distance}
We evaluate the surface following quality by measuring the distance between mesh surface and the obtained trajectories at $1$ cm intervals along the trajectory. 
\begin{figure}[h!]
    \centering
   \includegraphics[width=\columnwidth]{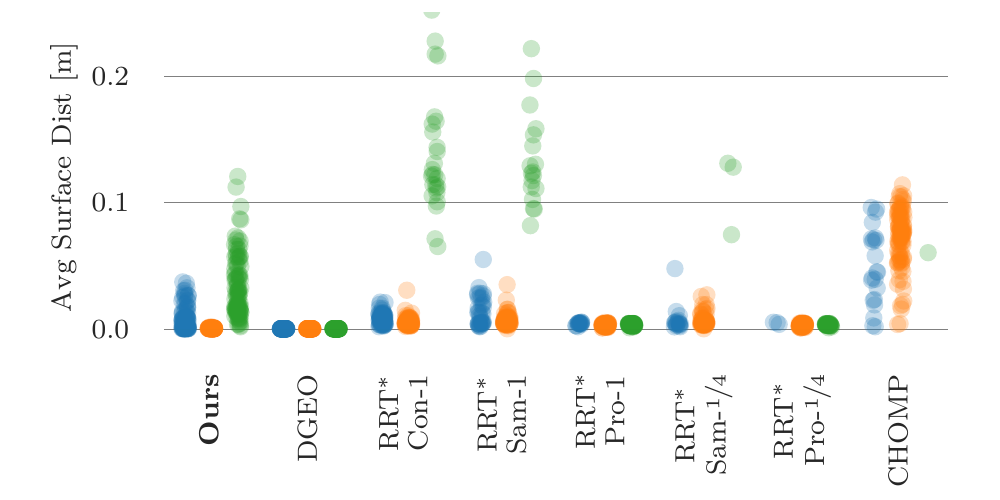}
   \includegraphics[width=\columnwidth]{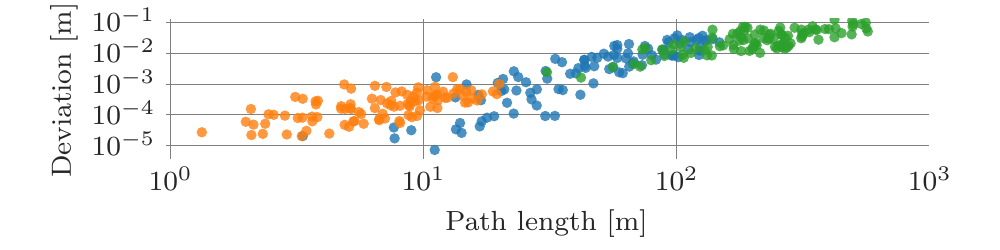}
     \vspace*{-8mm}
    \caption{Top: Average surface distance for each planned trajectory. Bottom: Scatter plot for all trajectories generated by the proposed planner that shows a correlation of path length and average surface distance deviation.}
    \label{fig:surface_distance}
    
\end{figure}
As shown in \cref{fig:surface_distance}, the proposed planner shows below $1$ mm deviation on average from the true surface for paths shorter than $10$ m and the \texttt{hilo} scenario. Deviations on the large \textit{rhone} mesh can be explained by the physical nature of the planner. Trajectories of multiple $100$ meters lead to larger velocities with the current tuning. Yet, there might be sharp changes in slope on the meshes, which the planner smooths to a certain extent as the trajectory is only affected by the resulting acceleration field. Depending on the use case, it can be advisable to adjust the planner tuning for very large maps.
The projection-based planners show less deviation, as they explicitly project the obtain trajectory onto the nearest surface. However, depending on the sampling quality and geometry this can lead to invalid paths as the projection might not be uniquely defined. The observed deviations of the CHOMP planner are in large parts due to the needed slack in the potential function.
Note that the plot only accounts for successfully planned trajectories and the success rate is relatively low on some scenarios for some planners (see \cref{fig:success_rate}).
\subsection{Path Length}
\begin{figure}[h!]
    \centering
    \includegraphics[width=\columnwidth]{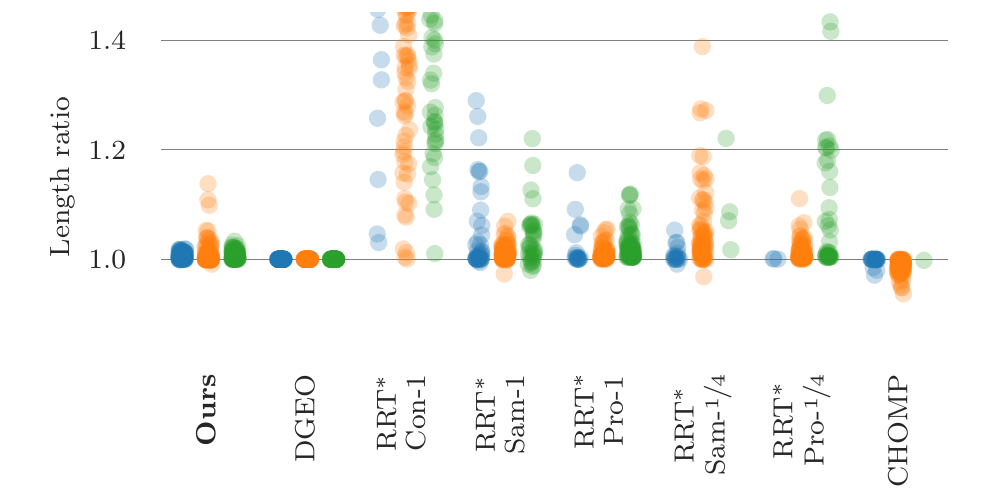}
    \vspace*{-8mm}
    \caption{Length ratios for all planners. The discrete geodesic algorithm is used as the benchmark and therefore its length ratios are by definition 1.0.}
    \label{fig:lengthratio}
\end{figure}

To show the optimality of the obtained paths as well as the effect of the induced distortion of $\mapping$ for the RMP planner, we compare trajectory lengths to the theoretical optimum obtained by the discrete geodesic algorithm.
The length ratio used here is defined as the trajectory length divided by the trajectory length for the same problem as obtained by the discrete geodesic algorithm. The closer to 1.0 the ratio, the closer to the theoretical optimum without any smoothing of abrupt edges or corners.
\Cref{fig:lengthratio} shows the results.
Except for RRT*-Con, most of the trajectories obtained by the RRT* variants are within reasonable bounds with a few outliers. However, as only successful plans are considered, there is a bias towards straight, simple paths. CHOMP tends to generate paths that are too short, i.e. intersect geometry.
The trajectories obtained with the proposed RMP-based planner are close to the optimum without outliers, effectively showing that the distortion induced by the mapping $\mapping$ seems of relatively small impact for the practical problems presented in this paper.
\subsection{Weighting of Policies}
\begin{figure}[!ht]
    \centering
    \includegraphics[width=\columnwidth]{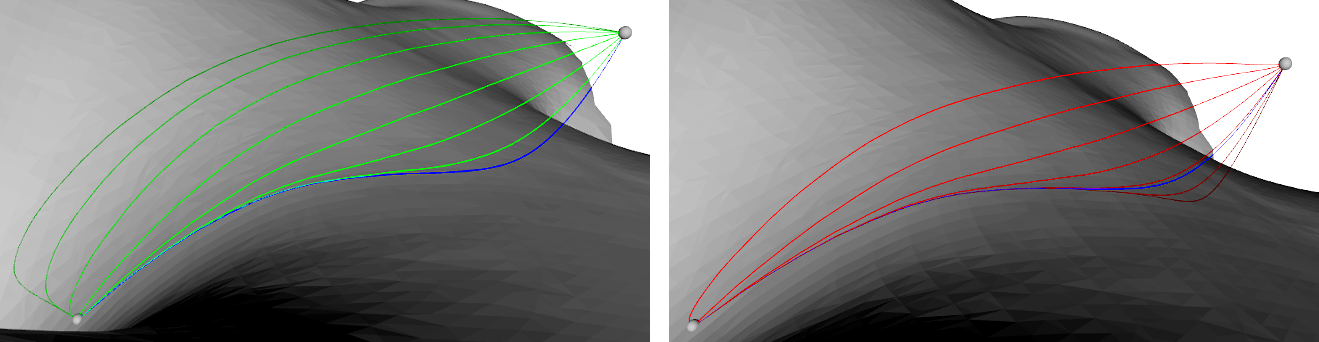}
    \vspace*{-4mm}
    \caption{Left: Variations of $\alpha^{\perp}$, where the darkest shade of green corresponds to a value of $0.1$ and the brightest to $25.6$. Right: Variations of $\alpha^{\surf}$ between $0.2$(darkest) and $25.6$. The green ball marks the start position, and the red ball the goal position. The blue trajectory corresponds to the tuning evaluated in the rest of the experiments.}
    \label{fig:weighting}
\end{figure}

The proposed planner is able to plan trajectories from free-space towards a goal on a surface. The behavior of this transition can be tuned by changing the $\alpha^{\perp}$ and $\alpha^{\surf}$ relative to each other. Intuitively, this can be interpreted as balancing the strength of the two policies relative to each other. \Cref{fig:weighting} shows the impact of the two parameters for a given planning problem.
As $\alpha^{\perp}$, respectively the strength of the surface attraction policy, approaches $0$ we obtain trajectories that stay at an approximately constant distance to the surface in curved parts, and exact constant distance in flat parts. Conversely, if the surface following policy is weaker relatively, the trajectory attaches to the surface as direct as possible.
As the planner generates an acceleration field that can be evaluated at each time step, the tuning can be adjusted mid-trajectory at any time.
\section{Conclusion}
\label{sec:conclusion}
A novel path planning framework that combines Riemannian motion planning with mesh manifolds has been presented in this paper.
The proposed framework solves the problem of approaching a surface and staying on a surface in 3D space in a mathematically elegant and real-world applicable way. The guarantees provided by the Riemannian motion policies combined with the proposed mesh-manifold rule out local-minima problems. Numerical inaccuracies could theoretically lead to non-optimal or non-terminating policies in rare cases, however we never observed such behavior in tests.
We showed that our approach outperforms others in terms of performance, robustness, and execution time and works well on dissimilar scenarios with the same tuning. Our proposed algorithm does not suffer from local-minima or tuning problems like optimization-based approaches and is not prone to sampling bottlenecks or geometrical ambiguity such as sampling-based planners.
The proposed planner possesses a range of very powerful properties that motivate many future directions of work.
The ability to follow a global near-optimum by the next best local direction at a very high rate allows the seamless integration of live sensor data to e.g. facilitate on-surface reactive and dynamic obstacle avoidance. Other directions include the addition of orientation policies and combination with learned policies based on surface properties.
\bibliography{mybib}{}
\bibliographystyle{IEEEtran}
\end{document}